\documentclass[12pt]{article}

\usepackage{booktabs}
\usepackage{amsmath}
\usepackage{amsthm}
\usepackage{amssymb}
\usepackage{cite}
\usepackage{url}
\usepackage[multiple]{footmisc}
\usepackage{graphicx}
\usepackage{epstopdf}
\usepackage{chngcntr}
\usepackage{enumitem}
\usepackage{hhline}
\usepackage{array}
\usepackage{hyperref}
\usepackage{color}
\usepackage{array}
\usepackage{multirow}
\usepackage{algorithm,algorithmic}
\usepackage[acronym]{glossaries}

\hypersetup{colorlinks=false}

\if@twoside
\else 
\fi

\numberwithin{equation}{section}
\counterwithin{figure}{section}
\counterwithin{table}{section}

\theoremstyle{plain}

\theoremstyle{definition}

\theoremstyle{remark}
\newtheorem{remark}{Remark}[section]

\definecolor{aog}{rgb}{0.0, 0.5, 0.0}

\makeatletter
\newcommand{\ALOOP}[1]{\ALC@it\algorithmicloop\ #1%
  \begin{ALC@loop}}
\newcommand{\ENDALOOP}{\end{ALC@loop}\ALC@it\algorithmicendloop}

\makeatother

\title{Leveraging Real-Time Data Analysis and Multiple Kernel Learning for Manufacturing of Innovative Steels}

\author{
Wolfgang Rannetbauer\footnote{voestalpine Stahl GmbH, voestalpine-Stra{\ss}e 3, A-4020 Linz, Austria (wolfgang.rannetbauer@voestalpine.at), Corresponding author.} ,
Simon Hubmer\footnote{Johannes Kepler University Linz, Institute of Industrial Mathematics, Altenbergerstra{\ss}e 69, A-4040 Linz, Austria, (simon.hubmer@jku.at)} ,
\\
Carina Hambrock\footnote{voestalpine Stahl GmbH, voestalpine-Stra{\ss}e 3, A-4020 Linz, Austria (carina.hambrock@voestalpine.com)} ,
Ronny Ramlau\footnote{Johannes Kepler University Linz, Institute of Industrial Mathematics, Altenbergerstra{\ss}e 69, A-4040 Linz, Austria, (ronny.ramlau@jku.at)} \footnote{Johann Radon Institute for Computational and Applied Mathematics, Altenbergerstra{\ss}e 69, A-4040 Linz, Austria, (ronny.ramlau@ricam.oeaw.ac.at)}
}

\begin{document}

\maketitle
\begin{abstract}

The implementation of thermally sprayed components in steel manufacturing presents challenges for production and plant maintenance. While enhancing performance through specialized surface properties, these components may encounter difficulties in meeting modified requirements due to standardization in the refurbishment process. This article proposes updating the established coating process for thermally spray coated components for steel manufacturing (TCCSM) by integrating real-time data analytics and predictive quality management. Two essential components--the data aggregator and the quality predictor--are designed through continuous process monitoring and the application of data-driven methodologies to meet the dynamic demands of the evolving steel landscape. The quality predictor is powered by the simple and effective multiple kernel learning strategy with the goal of realizing predictive quality. The data aggregator, designed with sensors, flow meters, and intelligent data processing for the thermal spray coating process, is proposed to facilitate real-time analytics. The performance of this combination was verified using small-scale tests that enabled not only the accurate prediction of coating quality based on the collected data but also proactive notification to the operator as soon as significant deviations are identified. 

\end{abstract}

\section{Introduction}
Over time, the production of conventional steel grades has evolved into a highly efficient and well-functioning process. A significant contribution to this progress is made by thermally spray coated components for steel manufacturing (TCCSM) \cite{ref18, matthews2010review}. The coating process for these components has developed alongside the steel manufacturing process and is now highly standardized. Depending on the application of the TCCSM, surface coatings are tailored to specific requirements, such as wear and corrosion resistance, which were similar when using traditional steels, thus ensuring efficient steel production \cite{ref18}.

However, in response to the global pursuit of sustainability and reduced CO\textsubscript{2} emissions, expectations on steel quality have changed considerably. The demand for innovative steels is illustrated by the ubiquity of advanced high-strength steels (AHSS) \cite{rimnac2019trends}. These find application in critical components such as battery housings and structural elements of electric vehicles, since they have a special synthesis of lightweight properties and exceptional fracture toughness \cite{singh2014review}. At the same time, the requirements on special steel variants, such as electrical steel with special magnetic properties, which are essential for wind turbine generators, are becoming increasingly stringent \cite{xia2008developments}. These modified quality issues in steel production represent a significant threat to TCCSM integrity for high-quality steel manufacturing \cite{matthews2010review, matthews2010review1}. The use of enhanced alloys in these innovative steels introduces entirely new stresses on TCCSM and thus requires modern and more adaptable coatings. Continuation of the established coating process carries the risk of failures and defects that either lead to minor quality cutbacks in the steel products or to massive costs due to production downtime and maintenance. The associated inefficiencies represent a significant obstacle to the efficient production of innovative steel products.

The rapid progress of data-driven technologies in the context of Industry 4.0 and Smart Manufacturing has led to a variety of innovative approaches and strategies \cite{ref6, ref7}, which offer a great opportunity to digitally transform the standardized coating process of TCCSM. One of them is known as real-time data analytics, enabled by the Industrial Internet of Things (IIoT). Referring to the literature \cite{ref7.5}, IIoT is defined as: an extension of network connectivity and computing capability to objects, devices, sensors, and items not ordinarily considered to be computers. This implies that knowledge and information gathered from humans, machinery, the environment, and the manufacturing process serve as an important foundation for real-time insights and informed decision-making \cite{ref8}. Towards a more customized and adaptable coating for TCCSM, accurate and reliable prediction of coating quality is essential. At the same time, this is also the largest hindrance, not least due to process variations during the coating process. Fortunately, the increasing popularity of machine learning (ML) in recent decades has yielded powerful and robust prediction models, such as support vector machines (SVM), and a variant known as multiple kernel learning (MKL) \cite{ref8.5, ref10, ref11}. Especially MKL is proven to have an excellent ability to make predictions for multivariate, potentially very heterogeneous data, whether for classification or regression tasks \cite{ref11.1, ref11.14, ref11.15, ref11.16, ref11.2}, making it a promising approach for a TCCSM coating quality predictor. Nevertheless, supervised ML algorithms require a number of labeled sample data effective for model training. However, in the field of thermal spraying, an area that is often characterized by manual processes, the availability of data in a suitable format is particularly limited. Data from existing machines can only be supplied if machines are replaced by new ones or retrofitted using smart devices \cite{ref12}. To overcome the challenge of data demand, a data aggregator needs to be implemented - an intelligent, downstream data management system capable of autonomously collecting relevant real-time data from the coating process through a robust IIoT infrastructure. Thus, the combination of a data aggregator and quality predictor appears to be an exceptional solution addressing the critical aspects of a modern and more adaptable TCCSM coating.

This article aims to explore one possibility to update the established coating process of TCCSM by a combination of predictive quality management and real-time data analytics. The contribution of this article can be summarized as follows.
\begin{enumerate}
    \item Application of a novel combination of continuous process monitoring and data-driven methods in the field of thermal spraying to facilitate and improve the coating procedure through their complementary cooperation.
    \item Concerning the data-driven methodology, a simple and efficient multiple kernel learning (SEMKL) model is developed, which uses pre-processed real-time data to predict the coating quality of TCCSM.
    \item With regard to continuous process monitoring, a data aggregator is proposed, utilizing sensors, flow meters, and intelligent data handling to enable real-time coating monitoring and analysis. This system proactively notifies operators when significant deviations in key quality parameters are detected, allowing for timely intervention.
\end{enumerate}  

This contribution is further distinguished by the practical implementation of both the Data Aggregator and Quality Predictor, which have been fully integrated and are operational within the production environment at voestalpine Stahl GmbH \cite{TSM}. This deployment moves beyond a conceptual framework, providing a validated, active system that continuously supports real-time quality management in thermal spraying operations.

\section{Related Works}

\subsection{MKL Prediction Models} \label{relatedMKL}
A central subgroup of ML models are kernel methods, in particular the variant of multiple kernel learning (MKL), which has proven to be remarkably efficient in various prediction tasks \cite{ref8.5, ref11.1, ref11.15, ref11.16, ref11.2}. Across several application domains, MKL has demonstrated strong performance in tackling prediction problems, including optimizing algorithm hyper-parameters in unsupervised domain adaptation \cite{ref_dinu}, forecasting tasks in signal processing applications \cite{ref13}, electric load forecasting \cite{ref11.2}, classification of hyperspectral images \cite{ref14}, fall detection \cite{ref15}, and applications in the steel industry like silicon content prediction in blast furnaces \cite{ref16}.

In addition to MKL, several other kernel-based techniques have been widely used in prediction tasks, such as Gaussian Process Regression (GPR), Kernel Ridge Regression (KRR), and Support Vector Machines (SVM). While each of these methods possesses distinct advantages, they also exhibit limitations that may render them less suitable for addressing the specific challenges associated with TCCSM quality prediction.

GPR offers probabilistic predictions and uncertainty quantification, which are advantageous in applications where uncertainty is critical \cite{refSVM, refGPR1, refGPR2}. However, since GPR relies on computationally intensive matrix inversions, which scale cubically with the number of data points, it is less efficient for larger datasets \cite{refGPR3}. Furthermore, GPR typically employs a single kernel, limiting its ability to handle multiple heterogeneous data sources, such as combining static features and diverse sensor data.

KRR, which integrates ridge regression with kernel methods, is powerful in capturing non-linear relationships within data. Nevertheless, a significant limitation of KRR lies in the challenge of selecting an appropriate kernel, a choice that influences model performance \cite{refSVM}. Traditional cross-validation techniques often encounter difficulties in identifying the optimal kernel, which limits the adaptability of KRR, especially when addressing heterogeneous data sources or complex patterns that require more flexible combinations of kernels \cite{refKRR1}.

SVM are renowned for their ability to perform classification and regression tasks in high-dimensional spaces through the identification of optimal hyperplanes \cite{refSVM}. However, similar to GPR and KRR, SVMs typically utilize a single kernel function, which restricts their ability to model complex relationships inherent in heterogeneous data sources \cite{ref10}.

In contrast, MKL enhances learning by simultaneously optimizing kernels and their weights, offering flexibility for handling diverse multivariate data. It assigns different weights to kernels based on their impact on outputs. MKL algorithms aim to find the optimal combination of kernels for effective learning. These algorithms can be divided into two categories: one-step methods and wrapper techniques. One-step algorithms employ advanced optimization methods and calculate both the kernel weights and the coefficients related to the learning problem in a single run. Wrapper methods address MKL by initially solving a single learning problem with a given set of kernel weights, followed by iterative weight updates. Such approaches usually run much faster than their one-step counterparts \cite{ref17}. 

Recent developments in learning paradigms have also highlighted the relevance of techniques like transfer learning (TL) for evolving traditional ML models. TL is particularly valuable when real process data is sparse, leveraging knowledge gained in a source domain to improve model performance in a target domain where data is limited \cite{refTL2}. In the context of TCCSM and coating quality prediction, TL could reduce the need for extensive real-world data collection by transferring insights from laboratory settings to operational manufacturing processes \cite{refTL}. This approach would not only lower costs but also address challenges related to operational constraints, such as safety and data availability.

Although MKL and wrapper methods are not inherently developed for transfer learning, their inherent flexibility in optimizing multiple kernels allows them to be adapted for certain transfer scenarios, particularly in environments where data collection is limited, such as certain manufacturing or maintenance processes. Wrapper methods have already proven useful during initial modelling phases in controlled environments \cite{ref11.1, ref11.2}. The knowledge gained from such models can serve as a basis for applying TL techniques in industrial-scale systems, thereby improving the adaptability of the MKL model and the overall performance.

Furthermore, continual learning (CL) has gained attention as a method for dealing with evolving data distributions \cite{refCL2}. Unlike traditional models that assume static data, CL allows models to adapt incrementally to new data while retaining knowledge from prior tasks \cite{refCL}. This is particularly relevant for industrial applications, where operational conditions and data evolve over time. While CL is beyond the immediate scope of this study, future work could explore the integration of MKL models into CL to maintain continuous adaptability in dynamic environments.

While MKL is commonly employed to address classification problems, this study applies the technique to predict the coating quality of TCCSM. To achieve real-time prediction, an efficient wrapper method known as Simple and Efficient Multiple Kernel Learning (SEMKL) is used in this paper. SEMKL provides a closed-form solution for updating the kernel weights, which significantly increases the computational efficiency \cite{ref11.1}. SEMKL extends the traditional SVM framework to multiple kernels, allowing for a more flexible modelling approach that can capture the heterogeneous nature of sensor data in TCCSM. By generalizing MKL to the $L_p$-norm of the kernel weights, SEMKL improves prediction accuracy through adaptive kernel weighting, ensuring that the most relevant features from diverse data sources contribute optimally to the prediction model. This flexibility is assumed to be crucial for TCCSM, where sensor data often varies in type and scale. Additionally, the computational efficiency of SEMKL, derived from its closed-form solution, makes it more suitable for real-time applications compared to similar techniques such as KRR and GPR, which face challenges with kernel selection and computational scalability. The promising results outlined in \cite{ref11.1} further support its use for complex prediction tasks like TCCSM quality forecasting.

Although SEMKL does not explicitly incorporate TL or CL, it lays the groundwork for future studies to explore these advanced learning techniques, particularly in scenarios where knowledge transfer from controlled laboratory environments to real-world industrial processes could provide considerable advantages.

\subsection{Coating Quality for TCCSM}
TCCSM are used across various phases of steel manufacturing, including steel making, continuous casting or coil rolling \cite{ref18}. As described in \cite{ref18}, the specific quality requirements for TCCSM vary according to their intended application, encompassing thermal insulation, wear, and corrosion resistance, and coatings with distinct chemical phase properties. Over time, standardized coating processes have been established for these components to meet their unique demands. While various studies have examined the optimization of coating quality properties for different applications, none have specifically addressed the dynamic requirements of TCCSM in response to evolving steel qualities. 

Prior research has explored the prediction and optimization of coating characteristics through methodologies such as Computational Fluid Dynamics (CFD) simulations \cite{ref19}, ML applications \cite{ref20}, and statistical models \cite{ref21, ref22, ref23, rannetbauer2024enhancing}. Common to these data-driven methods is their reliance on historical data and the assumption of a stable coating process. In practice, however, important input parameters such as the powder feed rate, gas flows, or pressures vary over the duration of the coating process, resulting in uncertainty in the predictions.

Inspired by the predictive capability of various coating properties shown in \cite{ref23}, another goal of predicting coating quality in this article is achieved from an MKL prospect. Moreover, in contrast to previous works, the quality prediction in this work relies on pre-processed real-time data generated by the data aggregator, aligning with a more dynamic and adaptable approach to coating quality prediction.

\section{Methodology}

\subsection{Conceptual Framework Overview}
\begin{figure*}[!ht]
\centerline{\includegraphics[width=\textwidth, height=8cm]{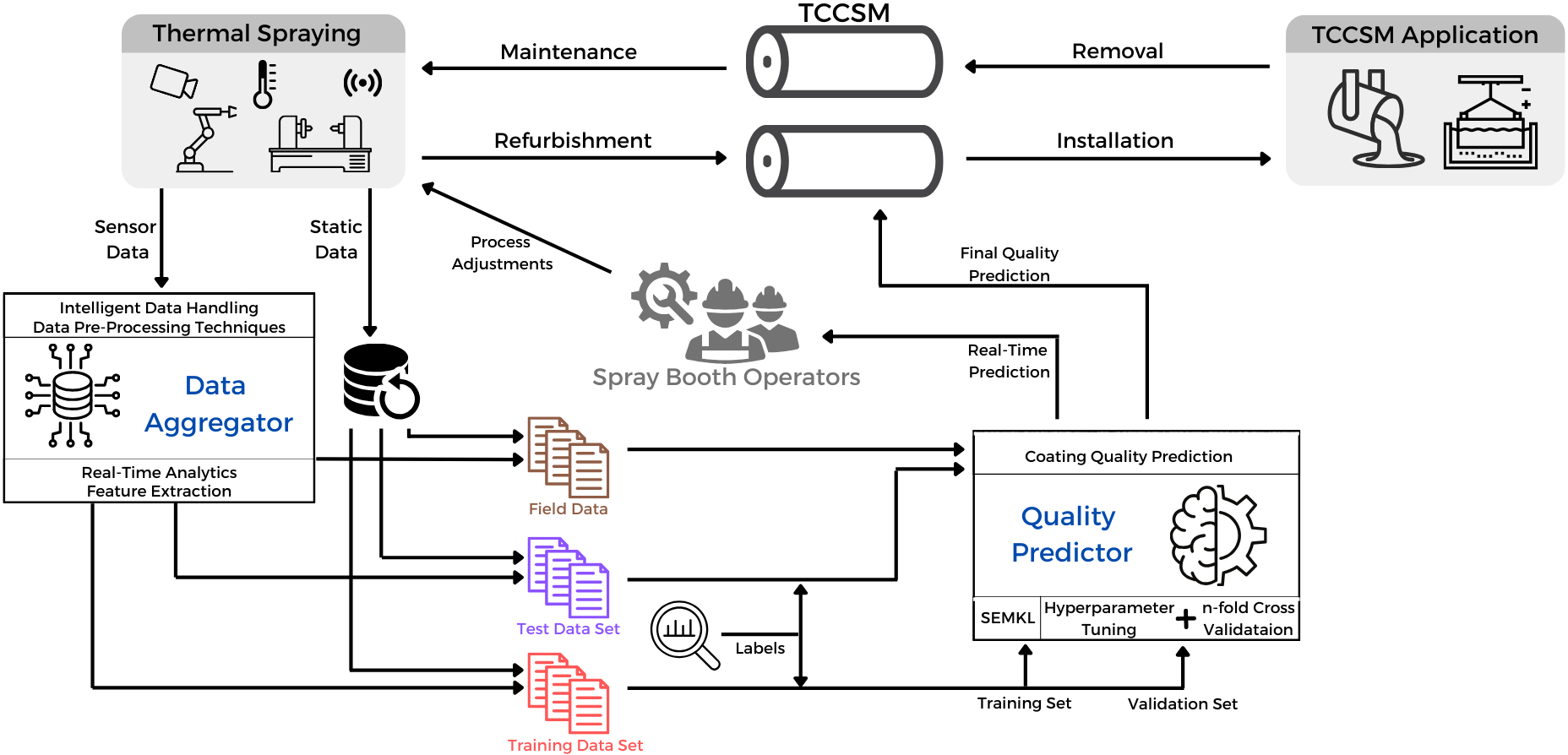}}
\caption{Framework of real-time analytics and quality prediction for thermally spray coated components in steel manufacturing (TCCSM).}
\label{fig:overview}
\end{figure*}

To provide clarity, Figure \ref{fig:overview} illustrates the conceptual framework adopted in this study. As described within the framework, two core components have been developed as outlined below. 

\begin{enumerate}
    \item Data Aggregator: The data aggregator serves as a sophisticated information technology system tailored to the real-time analysis of the thermal spray process. Referring to the literature \cite{refRTA}, real-time analysis is defined as the capability to perform computations and provide feedback within a minimal latency period after a triggering event. In the context of this study, real-time analysis refers to the ability to promptly process and analyze incoming sensor data during the coating process, allowing for the immediate detection of significant deviations from target process parameters, such as gas flow rates or system pressures. Through a complex network of sensors, flow meters, and advanced data processing functions, it not only independently collects and manages real-time data from the coating process, but also performs feature extraction. This process filters out relevant information and parameters from the sensor data, creating a customized data set suitable for training and testing ML models. Integrated into a robust IIoT infrastructure, the data aggregator ensures seamless data collection, enabling precise monitoring, real-time analysis, and the generation of curated data sets that are essential for the development and optimization of ML models.
    \item Quality Predictor: The Quality Predictor, based on SEMKL, accurately predicts coating quality for TCCSM. This predictor, trained, tuned, and validated on carefully prepared data sets from the data aggregator, is characterized by real-time predictions and enables thermal spray booth operators to make timely adjustments if an unsatisfactory expected coating quality is detected during the ongoing process. After coating, the predictor determines the final predicted quality of the TCCSM and guides the optimal use of components in the most suitable applications. This intelligent, data-driven methodology ensures proactive intervention and serves as a key part to ensure predictive quality.
\end{enumerate}

\subsection{Data Aggregator} \label{sect:agg}
Various factors have a significant influence on the thermal coating process, including the configuration of the gas flows, the system pressures, the temperatures, and their dynamic changes during the coating period. Moreover, the combination of these parameters, such as the fuel-oxygen ratio, is crucial and has a direct impact on the resulting coating quality \cite{ref22, ref23}. Achieving process stability is essential for optimal coating results. However, in industry, especially when coating large components such as TCCSM, the process gases vary throughout the coating process, which can take hours. These deviations from the target settings inevitably lead to variations in the coatings. To digitally capture the nuances of the coating process, sensors and other smart devices are used to accurately record all gas flows, pressures, temperatures, and other relevant process data.

\begin{enumerate}
    \item Intelligent Data Handling: The data aggregator operates through a programmable logic controller (PLC) embedded within the coating booth, enabling efficient recording and extraction of process data on a sequential basis. Since the coating process for TCCSM occurs intermittently--rather than continuously--the system monitors the operational status of the robot (whether it is in ``Coating'' mode or not) at a sampling rate of 250 ms. Once an active coating process is detected, the sampling rate for all relevant sensor data, such as gas flows, pressures, and temperatures, switches to 100 ms, ensuring high-frequency data capture during the active phases. Furthermore, the data is saved intelligently, with the initial value being recorded immediately. Subsequent values are only saved if a meaningful change occurs, e.g., change in compressed air flow is limited to deviations of $\pm1\%$ from the target value. This approach ensures that only data relevant to the coating process is recorded, optimizing both memory and processing efficiency. Figure~\ref{fig:medis} illustrates the intelligent data storage method for the recorded compressed air consumption during the coating process. On the left, a time series graph depicts the progression from the beginning to the end of the coating. The corresponding data table is shown on the right-hand side.

    \begin{figure}[!ht]
    \centerline{\includegraphics[width=\textwidth]{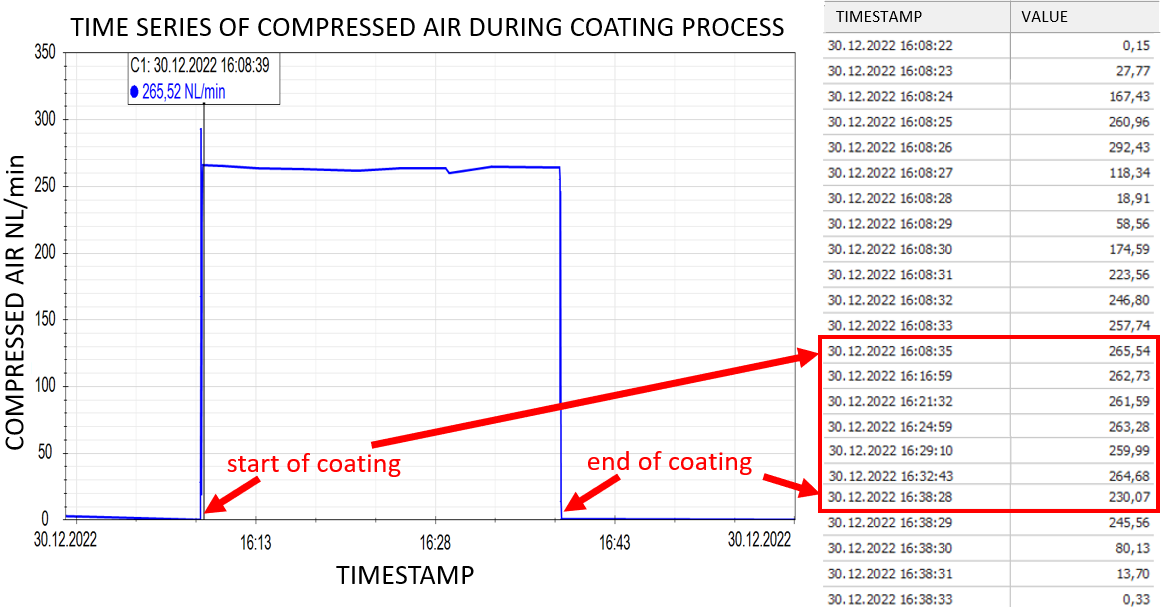}}
    \caption{Representation of the intelligent data storage for the recorded compressed air consumption during the coating process. The left-hand side shows the time series from the start to the end of the coating process, while the right-hand side shows the corresponding data table.}
    \label{fig:medis}
    \end{figure}

    \item Data Pre-Processing Techniques: In the context of temporal data recorded by different systems such as the controller, the robot, and pyrometer, timing is essential to synchronize different data streams. These systems utilize industrial network nodes (gateways) to bridge different transmission protocols, allowing data from diverse sources to be collected and unified before further analysis. To ensure a coherent and synchronized data set, a timestamp mechanism is used to ensure that the data points of the individual systems correspond to exactly the same point in time. This enables a reliable alignment of the streams despite potential differences in sampling rates or transmission delays.
    
    Missing values in sensor data due to technical breakdowns or temporary malfunctions are common in practice. For this particular application, the use of simple mean value imputation, based on the current coating process has proven to be effective and reliable. In the case that no data is recorded over the entire coating period, imputation using regression can be used, since at least some of the input parameters correlate with each other. 

    \item Real-Time Analytics: The data aggregator not only records and pre-processes data but also provides instant insights into the coating process. It identifies significant deviations or anomalies from specified target values, such as sudden withdrawals of compressed air due to parallel processes. Upon detecting such deviations, the system promptly displays the actual process variables through a web server and a proprietary application interface, enabling real-time monitoring. This includes both raw and calculated parameters, such as the fuel-oxygen ratio, ensuring that crucial metrics are available immediately.

    Live process visualization plays an important role in supporting operators. Historical data is graphically represented in line charts, providing a comprehensive overview of the entire coating sequence. The system’s real-time analysis is highly responsive, with an average time delay of 4.92 ms between event detection and display for non-calculated variables, as confirmed by server diagnostics. This allows operators to receive near-instant feedback on process status, enabling timely responses to any discrepancies from target settings. The continuous graphical visualization of the process, exemplified by the compressed air flow data in Figure~\ref{fig:medis} (left), further enhances operator oversight, providing a dynamic and accessible overview of the coating process. Operators can respond to system alerts by manually adjusting parameters, ensuring the coating process remains within target settings.

    \item Feature Extraction: Machine learning algorithms, such as MKL, require well-organized data for effective learning. The pre-processed sensor data needs to be further refined to attain a usable format. The data aggregator performs this task by extracting relevant features from the time series data and organizes them accordingly. Depending on the variable under consideration, different features are used. In the case of gas flows, this includes the weighted average (corresponding to the time stamp) and the standard deviation. For pyrometer data, on the other hand, features such as the component starting temperature, the maximum temperature of the component, and the heating rate are extracted. This feature extraction process is supported by collaboration with the in-house thermal spray technicians, who use their expertise to define the variables that potentially influence the resulting coating. The data aggregator systematically processes the pre-processed sensor data and extracts the defined features. These features, together with static data that depends on TCCSM parameters (e.g. coating material used), result in a useful data set that is suitable for training, validation, or testing of ML methods (cf. Figure~\ref{fig:overview}).
\end{enumerate}

\subsubsection{Data Processing Overview}
The data aggregator integrates several components to ensure efficient data capture and processing for the thermal coating process. Sensors installed within the system record gas flows, temperatures, and rates, with data acquisition managed by a programmable logic controller (PLC). The PLC dynamically adjusts the sampling rate based on operational status, ensuring high-frequency data capture during active coating periods.

Once collected, the data is transmitted using the OPC-UA protocol through industrial gateways, which unify different data transmission formats for further processing. Time synchronization across data streams is achieved via timestamping mechanisms, ensuring coherent alignment between controller, robot, and pyrometer data. Missing values are imputed using either mean value imputation or regression methods, depending on the scope of the data gap. The processed data is visualized in real-time on a graphical interface, providing operators with instant feedback on both raw and calculated parameters. Deviations from target values trigger visual alerts and warnings, allowing for immediate corrective actions. Data updates occur with a latency of 4.92 ms, ensuring near-instantaneous process monitoring.

Finally, important features are extracted from the time series data—such as weighted averages and standard deviations for gas flows—and structured into formats suitable for ML algorithms, such as MKL. This structured dataset enables further analysis, supporting process optimization and predictive modelling.

\subsection{Quality Predictor}
The TCCSM quality predictor uses a variant of multiple kernel learning introduced in \cite{ref11.1}, to predict quality-relevant properties, leveraging data supplied by the data aggregator. Multiple kernel learning (MKL) offers a sophisticated approach to handling heterogeneous data arising from coating processes. By leveraging multiple kernels, MKL effectively addresses the challenge of data dimensionality, enabling the representation of high-dimensional data in a lower-dimensional space \cite{ref11.2}. Additionally, the adaptability of MKL to diverse data modalities ensures robust analysis and insights extraction across a wide range of coating conditions \cite{ref11.15}. 

Let $X = \{(\mathbf{x}_n, y_n), n = 1,\dots,N\} \in \mathbb{R}^d \times \mathbb{R}$ be a data set of $N$-labeled training samples provided by the data aggregator. Here, the $d$-dimensional feature vector $\mathbf{x}_n$ represents the $n$-th sample and the label $y_n$ refers to the measured quality property of the $n$-th sample. The quality properties, i.e. labels, were assessed not only for the training data but also for the test data to ensure supervised learning and model evaluation. Kernel learning (KL) extends beyond classification tasks and addresses regression problems by using a carefully selected loss function for the formulation of a prediction model, denoted as $f(\mathbf{x})$. Choosing the well-known $\epsilon$-insensitive loss function $\ell_\epsilon$ defined in \cite{ref23.5}, where:
\begin{align}
    \ell_\epsilon(f(\mathbf{x}), y) = \begin{cases}
0, \quad\quad\quad\quad\quad\quad & \text{if } |f(\mathbf{x}) - y| < \epsilon,\\
|f(\mathbf{x}) - y| - \epsilon, \quad & \text{otherwise,}
\end{cases}
\end{align} 
for some fixed $\epsilon \geq 0$, the regression model $f(\mathbf{x})$ using KL is then of the (dual) form \cite[Chapter 7.1.4]{ref8.5}
\begin{align} \label{model}
   f(\mathbf{x}) = \sum_{n=1}^N (\alpha_n - \alpha_n^*) K(\mathbf{x},\mathbf{x}_n) + b,
\end{align}
where $\alpha_n, \alpha_n^*$ denote the optimized support vectors (= Lagrange multipliers) and $b$ denote the bias term. These coefficients are learned from the data $X$. The $\epsilon$-insensitive loss function ensures robustness against minor deviations in the data, leading to a sparse solution \cite{ref8.5}, and aligns with established best practices in support vector regression. In the case of MKL regression the kernel function $K(\mathbf{x},\mathbf{x_n})$ is basically a linear combination of basis kernels 
\begin{align} \label{MKL}
   K(\mathbf{x},\mathbf{x}_n) = K(\mathbf{x},\mathbf{x}_n;\boldsymbol{\gamma}) = \sum_{m=1}^{M}\gamma_m K_m(\mathbf{x},\mathbf{x}_n),
\end{align}
where $M$ symbolizes the total number of basis kernels $K_m:~\mathbb{R}^d\times \mathbb{R}^d \rightarrow \mathbb{R}$ for $m \in \{1,\dots,M\}$, and $\boldsymbol{\gamma} = (\gamma_m)_{m=1}^M$ are the respective kernel weights. This approach is particularly suited for capturing the diverse characteristics of coating quality data, as it allows for the integration of multiple kernels, offering greater flexibility and accuracy compared to single-kernel methods. As explained in Section~\ref{relatedMKL}, the MKL problem involves the simultaneous optimization of coefficients ($\alpha$, $\alpha^*$, and $b$) and kernel weights $\gamma_m$ within a unified optimization framework. Using the loss-function $\ell_\epsilon$, the objective of MKL regression is to learn a prediction model $f(\mathbf{x})$, which is a linear combination of $M$ kernels, by solving the following optimization problem \cite{ref11.1}:
\begin{align} \label{eq4}
   \min_{\gamma \in \Delta} \min_{f \in \mathcal{H}_\gamma} \frac{1}{2} ||f||^2_{H_\gamma} + C \sum_{n=1}^{N} \ell_\epsilon(f(\mathbf{x}_n),y_n),
\end{align}
where $\mathcal{H}_\gamma$ is a reproducing kernel Hilbert space (RKHS) induced by the kernel function $K(\mathbf{x},\mathbf{x}_n;\boldsymbol{\gamma})$ and $C > 0 \in \mathbb{R}$ is a predefined positive regularization parameter, which controls the trade-off between model simplicity and regression error. Choosing a natural domain for $\boldsymbol{\gamma}$, denoted as $\Delta = \{ \boldsymbol{\gamma} \in \mathbb{R}^M_+: \sum_{m=1}^M \gamma_m=1, \gamma_m \geq 0\}$, results in a sparse configuration of kernel weights, characterized by an $l_1$-norm. However, it is important to note that this sparsity may lead to suboptimal performance, as discussed in \cite{ref24}. Following the suggestions in \cite{ref11.1}, we choose a more general domain of $\boldsymbol{\gamma}$, i.e., $\Delta = \{ \boldsymbol{\gamma} \in \mathbb{R}^M_+: ||\boldsymbol{\gamma}||_p \leq 1, \gamma_m \geq 0\}$, where $||\cdot||_p$ is the $l_p$-norm of the kernel weights, defined by $||\boldsymbol{\gamma}||_p = \big( \sum_{m=1}^M |\gamma_m|^p \big)^\frac{1}{p}$.

\begin{remark}
    The existence of a reproducing kernel Hilbert space (RKHS) depends on the positive definiteness of the kernel $K(\mathbf{x},\mathbf{x}_n;\boldsymbol{\gamma})$ \cite[Theorem 1.2]{ref17}. However, allowing negative values in $\boldsymbol{\gamma}$, can result in a loss of positive definiteness, potentially invalidating the existence of an RKHS. Although negative weights $\gamma_m$ may introduce greater flexibility, they can lead to methodological violations, rendering the learning problem ill-posed. Specifically, the kernel function may become indefinite, introducing numerical instabilities and leading to unreliable or non-convergent solutions during the optimization process \cite{refRKHS}.
    
    To avoid these issues, we restrict the kernel weights $\gamma_m \geq 0$, ensuring that the resulting kernel is a positive linear combination of base kernels, each of which is assumed to be positive definite. This constraint preserves the positive definiteness of the overall kernel $K(\mathbf{x},\mathbf{x}_n;\boldsymbol{\gamma})$, thereby guaranteeing the existence of a valid RKHS and maintaining the well-posedness of the learning problem \cite{ref8.5}.
\end{remark}

The hyperparameter $p \geq 1$ which yields the best generalization performance will be determined by cross validation. Thus, in this study, we consider the formulation of the (primal) MKL regression problem:
\begin{align} \begin{split} \label{primalMKL}
    \min_{||\boldsymbol{\gamma}||_p\leq1} \underset{\xi_n, \xi_n^* \geq 0}{\min_{\{ f_{\gamma_m} \in \mathcal{H}_{\gamma_m} \}_{m=1}^M}} \frac{1}{2} \sum_{m=1}^M \frac{1}{\gamma_m} ||f_{\gamma_m}||_{\mathcal{H}_{\gamma_m}}^2 + \ C \sum_{n=1}^{N} \bigg( \xi_n+\xi_n^* \bigg) , 
\end{split}\end{align} 
\begin{align} \text{s.t.} \ \ \ \ \ &\xi_n, \xi_n^* \geq 0, \quad \forall \ n \in \{ 1,\dots, N\}, \nonumber \\ 
  &y_n \leq \sum_{m=1}^M f_m(\mathbf{x}_n) + \epsilon + \xi_n, \quad \forall \ n \in \{ 1,\dots, N\}, \nonumber \\
  &y_n \geq \sum_{m=1}^M f_m(\mathbf{x}_n) - \epsilon - \xi_n^*, \quad  \forall \ n \in \{ 1,\dots, N\}, \nonumber \\
  & ||\boldsymbol{\gamma}||_p\leq1, \quad \gamma_m \geq 0, \quad  \forall \ m \in \{ 1,\dots, M\}, \nonumber
\end{align}
where $\xi_n$ and $\xi_n^*$ are the slack variables corresponding to the $n$-th training instance, and $\mathcal{H}_{\gamma_m}$ is, as before, the RKHS associated with $K(\mathbf{x},\mathbf{x}_n;\gamma_m)$. 

\begin{remark} 
    To transform the $\epsilon$-insensitive loss function $\ell_\epsilon$ into a constrained optimization problem, two slack variables $\xi_n \geq 0$ and $\xi_n^* \geq 0$ are introduced for each data point to account for prediction errors exceeding the threshold $\epsilon$ \cite{ref8.5}. Specifically, the conditions:
    \[
    \xi_n > 0 \quad \Leftrightarrow \quad y_n > f(\mathbf{x}_n) + \epsilon, \quad \text{and} \quad \xi_n^* > 0 \quad \Leftrightarrow \quad y_n < f(\mathbf{x}_n) - \epsilon,
    \]
    indicate that these slack variables penalize prediction errors exceeding the threshold $\epsilon$. The original $\epsilon$-insensitive loss function $\sum_{n=1}^{N} \ell_\epsilon(f(\mathbf{x}_n),y_n)$ in \eqref{eq4} is thus re-expressed in terms of the slack variables, leading to the objective function in equation \eqref{primalMKL}, which incorporates the penalization of these deviations through the term $\sum_{n=1}^N (\xi_n + \xi_n^*)$.
\end{remark}

Through this reformulation with slack variables $\xi_n$ and $\xi_n^*$, the primal MKL regression problem \eqref{primalMKL} can be transformed into the following dual representation, which is based on the standard support vector regression formulation (see \cite{ref8.5} for details) 
\begin{align} \begin{split} \label{dualMKL}
    \min_{||\boldsymbol{\gamma}||_p\leq1} \max_{\alpha, \alpha^* \in [0,C]^n} \sum_{n=1}^N (\alpha_n - \alpha_n^*) y_n \ - \epsilon \sum_{n=1}^N (\alpha_n - \alpha_n^*) \\ - \ \frac{1}{2} \sum_{n=1}^N \sum_{i=1}^N (\alpha_n - \alpha_n^*)(\alpha_i - \alpha_i^*) K(\mathbf{x},\mathbf{x}_n;\gamma) , 
\end{split}\end{align} 
\begin{align} \text{s.t.} \ \ \ \ \ & \sum_{n=1}^N (\alpha_n - \alpha_n^*) = 0, \nonumber \\
  & 0 \leq \alpha_n, \alpha_n^* \leq C, \quad \forall \ n \in \{ 1,\dots, N\}, \nonumber \\ 
  & ||\boldsymbol{\gamma}||_p\leq1, \quad \gamma_m \geq 0, \quad \forall \ m \in \{ 1,\dots, M\}. \nonumber
\end{align}
With $\boldsymbol{\gamma}$ constant, the maximization problem involving $\alpha$ and $\alpha^*$ is resolved using a standard support vector regression solver, such as the Sequential Minimal Optimization (SMO) algorithm outlined in \cite{ref25}. This algorithm decomposes a quadratic programming problem into a series of smaller, analytically solvable problems. Subsequently, when $\alpha$ and $\alpha^*$ are fixed at the optimized values in the min-max problem \eqref{dualMKL}, the remaining task of updating the kernel weights $\boldsymbol{\gamma}$ is tackled using a closed-form solution known as SEMKL \cite{ref11.1},

\begin{align} \label{update}
    \gamma_m = || f_{\gamma_m} ||_{\mathcal{H}_{\gamma_m}}^\frac{2}{1+p} \Bigg/ \bigg( \sum_{k=1}^M || f_{\gamma_k} ||_{\mathcal{H}_{\gamma_k}}^{\frac{2p}{1+p}} \bigg)^{\frac{1}{p}}.
\end{align}
These two steps keep iterating until a convergence criteria is reached, i.e., either the duality gap condition, defined in \cite{ref11}, or a maximum number of iterations is reached. Algorithm \ref{alg:SEMKL} provides a summary of the optimization process for the employed wrapper method in predicting the quality of TCCSM. 

\begin{algorithm} 
 \caption{The $l_p$ SEMKL regression }
 \begin{algorithmic}[1]\label{alg:SEMKL}
  \STATE Initialize $\gamma^0 = 1/M$
  \REPEAT 
  \STATE Solve the maximization part of the dual problem \eqref{dualMKL} with $K(\mathbf{x},\mathbf{x}_n;\boldsymbol{\gamma}) = \sum_{m=1}^{M}\gamma_m K_m(\mathbf{x},\mathbf{x}_n)$ to obtain an optimal solution for $\alpha$ and $\alpha^*$
  \STATE Calculate $|| f_{\gamma_m} ||_{\mathcal{H}_{\gamma_m}}^\frac{2}{1+p}$ and $|| f_{\gamma_k} ||_{\mathcal{H}_{\gamma_k}}^\frac{2p}{1+p}$
  \STATE Calculate $\gamma_m$ according to \eqref{update}
  \UNTIL Convergence
 \end{algorithmic} 
 \end{algorithm}

After the training process, the derived model $f(\mathbf{x})$ follows the structure defined by equation \eqref{model}, using the kernel function specified in equation \eqref{MKL}. Given a new feature vector $\mathbf{x}_{new}$ the model uses the learned coefficients $\alpha$ and $\alpha^*$, as well as the optimized kernel weights $\boldsymbol{\gamma}$, to generate predictions based on the formulated regression model $f(\mathbf{x})$.

\section{Case Study}
To explore the practical application of the proposed framework, a case study is conducted. Due to the challenge of acquiring extensive TCCSM data on a large scale within a reasonable time frame--given that the integration of components into diverse applications typically extends over several months or even up to a year--small-scale tests are employed. Although it would be ideal to conduct direct analyses on TCCSM data, the prolonged application durations make this impractical. Consequently, these small-scale tests, designed to cover a broad range of technical process capabilities, serve as a useful alternative, providing sufficient data for MKL model training and validation. This framework is currently operational within the production environment at voestalpine Stahl GmbH and is effectively integrated into the TCCSM coating processes. The training set consists of $N=49$ samples, each representing a specific small-scale experiment conducted to assess the coating quality of TCCSM. The same set is used for validation through $N$-fold cross-validation (CV), which allows for model tuning and performance evaluation on different subsets of the training data. Additionally, a separate test set comprising 10 samples is used to evaluate the generalization performance of the proposed framework on unseen data. 

The small-scale experiments are systematically designed to explore variations in coating conditions, with each experiment reflecting a distinct configuration of input parameters across a spectrum of process capabilities \cite{ref23}. Although sensor data from the coating process is time-dependent, datasets from large-scale components are difficult to collect in this context due to the prolonged application durations and integration of components across diverse industrial settings. Given the infrequent occurrence of TCCSM coatings, the availability of time-dependent data remains limited, as coatings occur intermittently. To facilitate data collection within a practical timeframe, the experiments were conducted on a smaller scale, with reduced coating time. While the component size in these tests differs from TCCSM, the coating conditions remain representative \cite{ref21}. Though the dataset comprises 49 training samples, obtaining these required coating operations conducted over two shifts daily for a week, using approximately 40 kg of coating material at an expense of about €120/kg. Destructive analysis of the samples followed, with results available approximately two months later, highlighting the time-intensive and resource-demanding nature of data generation in this context.

The time-dependent data is processed by the data aggregator, which transforms raw sensor readings into meaningful features suitable for machine learning algorithms. Details of these features are provided in the next paragraph and summarized in Table~\ref{tab:features}. This approach was developed in collaboration with coating engineers to ensure that the extracted features accurately represent the key aspects of TCCSM quality and performance. Several properties are critical when assessing the coating quality of TCCSM. Table~\ref{tab:properties} outlines these essential characteristics, categorizing them into ``Particle In-Flight Properties'' (PIP), ``Process Performance Properties'' (PPP), and ``Coating Quality Properties'' (CQP). 

\begin{table}[!ht]
\caption{Important properties for TCCSM coating quality assessment}
\centering
\footnotesize
\setlength{\tabcolsep}{3pt}
\begin{tabular}{p{140pt}p{150pt}p{130pt}}
\hline\hline
\vfil \textbf{Particle In-Flight Prop.} & \vfil \textbf{Process Performance Prop.} & \vfil \textbf{Coating Quality Prop.}  \\
\hline
Particle Velocity & Deposition Rate & Coating Thickness \\
Particle Temperature & Deposition Efficiency & Coating Roughness \\
 &  & Surface Hardness \\ 
 &  & Coating Porosity \\
\hline\hline
\end{tabular}
\label{tab:properties}
\end{table}

The ``Particle In-Flight Properties'' offer insights into the entire spraying process, serving as an initial indicator of process stability. Meanwhile, the ``Process Performance Properties'' encompass economically significant factors like deposition efficiency, contributing significantly to sustainable production. Finally, the ``Coating Quality Properties'' cover mechanical properties important for evaluating TCCSM quality, including surface hardness, coating thickness, roughness, and porosity.

A set of 27 features, listed in Table \ref{tab:features}, was extracted from each conducted small-scale experiment using the data aggregator. Predominantly, the data aggregator provides statistical features, encompassing parameters like time-weighted average gas flows or pressures. Additionally, technically relevant features such as the fuel-oxygen ratio are included. The choice of features for prediction by the quality predictor—whether pertaining to PIP, PPP, or CQP—varies based on the specific quality-relevant aspect and is indicated in the table. An underlying assumption, verified by in-house thermal spraying experts, suggests that environmental conditions predominantly affect CQP, while component-specific attributes, such as the heating rate, have negligible influence on the PIP.

\begin{table}[!ht]
\caption{Summary of 27 features extracted by the data aggregator from sensor data and their potential influence on PIP, PPP, and CQP}
\centering
\footnotesize
\setlength{\tabcolsep}{3pt}
\begin{tabular}{p{30pt}p{300pt}p{25pt}p{25pt}p{25pt}}
\hline\hline
\vfil No.& \vfil Feature Name & \vfil\hfil PIP & \vfil\hfil PPP & \vfil\hfil CQP \\
\hline
\vfil 1-3 & Weigh. Average of Environmental Conditions \par (Air Pressure, Humidity, Temperature) &  &  & \vfil\hfil \checkmark \\
4 & Defined Stand-off distance & \hfil \checkmark & \hfil \checkmark & \hfil \checkmark \\
5 & Defined Coating Velocity &  & \hfil \checkmark & \hfil \checkmark \\
6 & Defined Powder Feed Rate & \hfil \checkmark & \hfil \checkmark & \hfil \checkmark \\
7 & Weigh. Average TCCSM Cooling Temperature &  & \hfil \checkmark & \hfil \checkmark \\
8 & Weigh. Average TCCSM Cooling Flow Rate &  & \hfil \checkmark & \hfil \checkmark \\
\vfil 9-11 & Weigh. Average Process Gas Flow Rate \par (Fuel Gas, Oxygen, Shroud Gas) & \vfil \hfil \checkmark & \vfil \hfil \checkmark & \vfil \hfil \checkmark \\
\vfil12-17 & Weigh. Average Process Gas Inlet and Outlet Pressures (Fuel Gas, Oxygen, Shroud Gas) & \vfil\hfil \checkmark & \vfil\hfil \checkmark & \vfil\hfil \checkmark \\
18 & Weigh. Average Propane Temperature & \hfil \checkmark & \hfil \checkmark & \hfil \checkmark \\
19 & Weigh. Average AirJet Cooling Flow Rate & \hfil \checkmark & \hfil \checkmark & \hfil \checkmark \\
20 & Calculated Fuel-Oxygen Ratio ($\lambda$) & \hfil \checkmark & \hfil \checkmark & \hfil \checkmark \\
21 & Maximum TCCSM Temperature &  & \hfil \checkmark & \hfil \checkmark \\
22 & TCCSM Heat-up Rate &  & \hfil \checkmark & \hfil \checkmark \\
23 & TCCSM Cool-down Rate &  & \hfil \checkmark & \hfil \checkmark \\
24 & TCCSM Starting Temperature &  & \hfil \checkmark & \hfil \checkmark \\
\vfil 25-27 & Standard Deviation of Process Gas Flow Rate \par (Fuel Gas, Oxygen, Shroud Gas) &  & \vfil \hfil \checkmark & \vfil \hfil \checkmark \\
\hline\hline 
\multicolumn{5}{p{430pt}}{The check marks indicate whether the feature was used to predict the respective properties.}\\
\multicolumn{5}{p{380pt}}{\vfil PIP = Particle In-flight Properties,  PPP = Process Performance Properties, CQP = Coating Quality Properties }
\end{tabular}
\label{tab:features}
\end{table}

Candidate kernels for the SEMKL quality predictor are: a linear, second, and third degree polynomial kernel, and Gaussian kernels with seven distinct $\sigma^2$ values (0.05, 0.10, \dots, 0.35), resulting in a total of $M=10$ base kernels. In the model selection process, $N$-fold cross-validation (CV) was employed to assess the performance of the SEMKL models, involving variations in the regularization parameter $C$ and the hyperparameter $p$ representing the $l_p$-norm of kernel weights. Specifically, for each combination of $C$ and $p$, the training set was iteratively split, leaving one sample out for validation and using the remaining $N-1$ samples for training. This process was repeated for each sample, ensuring comprehensive validation with a total of $N = 49$ predictions. The $\epsilon$ parameter was set to a constant default value of 0.1 during this selection process, as exploratory tests indicated that deviations from this value provided no substantial improvement in predictive performance. Since model accuracy was observed to be more sensitive to variations in $C$ and $p$, the tuning process primarily focused on optimizing these parameters. The evaluation metric used to determine the performance of each SEMKL model $f(\mathbf{x})$ was the root mean squared deviation (RMSD), computed using the following formula:
\begin{align} \label{RSMD}
    RMSD(f(\mathbf{x})) = \sqrt{\frac{1}{N} \sum_{i=1}^N(y_i-f(\mathbf{x}_i))^2}
\end{align}
where $y_i$ represents the actual value of the considered quality property, $f(\mathbf{x}_i)$ denotes the predicted value, and $N$ is the total number of predictions. The choice of RMSD as the evaluation metric is attributed to its ability to provide a comprehensive measure of predictive performance by accounting for all deviations, irrespective of their magnitude. This differs from the $\ell_\epsilon$ loss, which introduces insensitivity to errors below a specified threshold, $\epsilon = 0.1$. While using a smaller $\epsilon$ may mitigate this insensitivity, RMSD offers better interpretability by uniformly considering all deviations and penalizing larger errors more heavily, facilitating a clearer understanding of model performance and enabling direct comparisons with baseline models.

The performance of the SEMKL model for describing particle in-flight velocity under various hyperparameters $C$ and $p$ is detailed in Table \ref{tab:RSMD}. Notably, the optimal SEMKL model configuration is attained with $C=10^4$ and $p=2^{10}$, as evidenced by the lowest RMSD value of 8.88.  In contrast, the RMSD value achieved by the standard linear model (LM) is significantly higher, i.e., 22.24 for predicting particle in-flight velocity. Consistently across all quality attributes considered, increasing the hyperparameter $p$ initially leads to improved prediction accuracy, which is in agreement with the results reported in \cite{ref11.1}. However, above a certain threshold, typically around $p=2^{9}$ or $p=2^{10}$, the accuracy tends to peak, indicating that the performance decreases with further increases in $p$. 

\begin{remark}
    Although values of $p=1$ and $p=2$ yielded satisfactory results, they did not achieve minimal error, as shown in Table \ref{tab:RSMD}. While the values of $p=2^{9}$ and $p=2^{10}$ in the $l_p$-norm are unconventional, they are consistently selected as optimal across all properties in the cross-validation process, leading to improved prediction accuracy. This aligns with observations in the original SEMKL study \cite{ref11.1}, where unusual high values for $p$ were found to enhance prediction accuracy, cf. \cite[Table 3]{ref11.1}. Additionally, the RMSD values observed for the test set, detailed in Table~\ref{tab:comparisionLM}, further demonstrate the effectiveness of the SEMKL approach, particularly in comparison to the higher error values observed with a classical linear model.
\end{remark}

The observed decline in performance for higher $p$, particularly beyond $p = 2^{10}$, can be attributed to numerical issues inherent to the $||\boldsymbol{\gamma}||_p\leq1$ constraint. For such high values of $p$, the $l_p$-norm behavior closely approximates the $l_\infty$-norm resulting in a situation where $|\gamma_m|^p$ tends to either 0 or infinity. This leads to numerical instabilities, which may affect the optimization process. Moreover, with increasing $p$, the SEMKL algorithm tends to converge faster, primarily due to the duality gap condition being satisfied early in the training phase. While faster convergence can be beneficial, it may inadvertently shorten the training process, potentially leading to suboptimal solutions and thus poorer performance for larger $p$. This reveals a practical limit for $p$, with peak performance consistently observed around $p=2^{9}$ or $p=2^{10}$ for all considered coating characteristics. Further exploration of alternative convergence criteria may be necessary to mitigate the impact of early convergence at higher values for $p$.

With an increase in $C$, the model performance for PIP and PPP kept improving until a certain point, beyond which further increments did not significantly enhance performance. For CQP, optimal performance is observed with lower values of $C$ ranging from 1 to 100.

\begin{table}[!ht]
\caption{RMSD values representing the performance of the SEMKL model across various combinations of hyperparameters $C$ and $p$ for characterizing particle in-flight velocity.}
\centering
\footnotesize
\setlength{\tabcolsep}{3pt}
\begin{tabular}{l|cccccc}
  \hline  \hline
 & $C=10^0$ & $C=10^1$ & $C=10^2$ & $C=10^3$ & $C=10^4$ & $C=10^5$ \\ 
  \hline
$p=2^0$ & 15.92 & 15.49 & 13.48 & 12.98 & 10.46 & 10.45 \\ 
$p=2^1$ & 15.85 & 15.41 & 12.98 & 12.23 & 10.40 & 10.39 \\ 
$p=2^2$ & 15.94 & 15.05 & 12.60 & 11.55 & 10.26 & 10.37 \\ 
$p=2^3$ & 15.92 & 14.75 & 12.47 & 11.34 & 10.41 & 10.21 \\ 
$p=2^4$ & 15.75 & 14.73 & 12.34 & 11.06 & 10.21 & 10.23 \\ 
$p=2^5$ & 15.75 & 14.45 & 12.39 & 11.05 & 10.23 & 10.26 \\ 
$p=2^6$ & 15.73 & 14.31 & 12.40 & 10.87 & 9.91 & 10.02 \\ 
$p=2^7$ & 15.73 & 14.28 & 12.45 & 10.92 & 9.64 & 9.90 \\ 
$p=2^8$ & 15.72 & 14.20 & 12.39 & 10.98 & 9.45 & 9.92 \\ 
$p=2^9$ & 15.72 & 14.22 & 12.39 & 10.81 & 9.07 & 9.82 \\ 
$p=2^{10}$ & 15.73 & 14.14 & 12.41 & 10.78 & \textbf{8.88} & 9.86 \\ 
$p=2^{11}$ & 15.78 & 14.17 & 12.52 & 10.87 & 9.02 & 9.88 \\ 
$p=2^{12}$ & 15.77 & 14.19 & 12.55 & 10.90 & 9.07 & 9.96 \\ 
$p=2^{13}$ & 15.81 & 14.24 & 12.56 & 10.93 & 9.14 & 10.05 \\ 
$p=2^{14}$ & 15.86 & 14.26 & 12.61 & 10.98 & 9.22 & 10.17 \\ 
$p=2^{15}$ & 15.97 & 14.34 & 12.69 & 11.03 & 9.41 & 10.33 \\ 
  \hline  \hline
\end{tabular}
\label{tab:RSMD}
\end{table}

\begin{figure}[!ht]
\centerline{\includegraphics[height= 8cm]{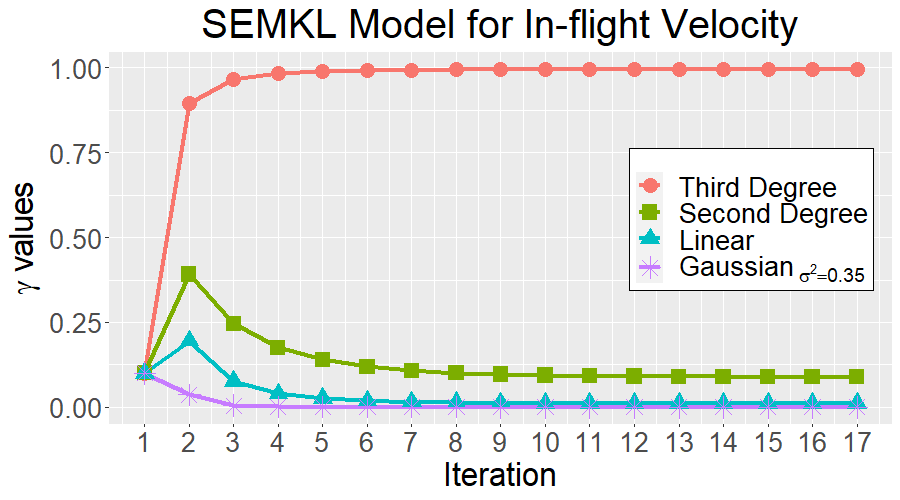}}
\caption{Evolution of the kernel weight values in SEMKL model for predicting particle in-flight velocity.}
\label{fig:kernel_weights}
\end{figure}

\begin{figure*}[!ht]
\centerline{\includegraphics[width=\textwidth]{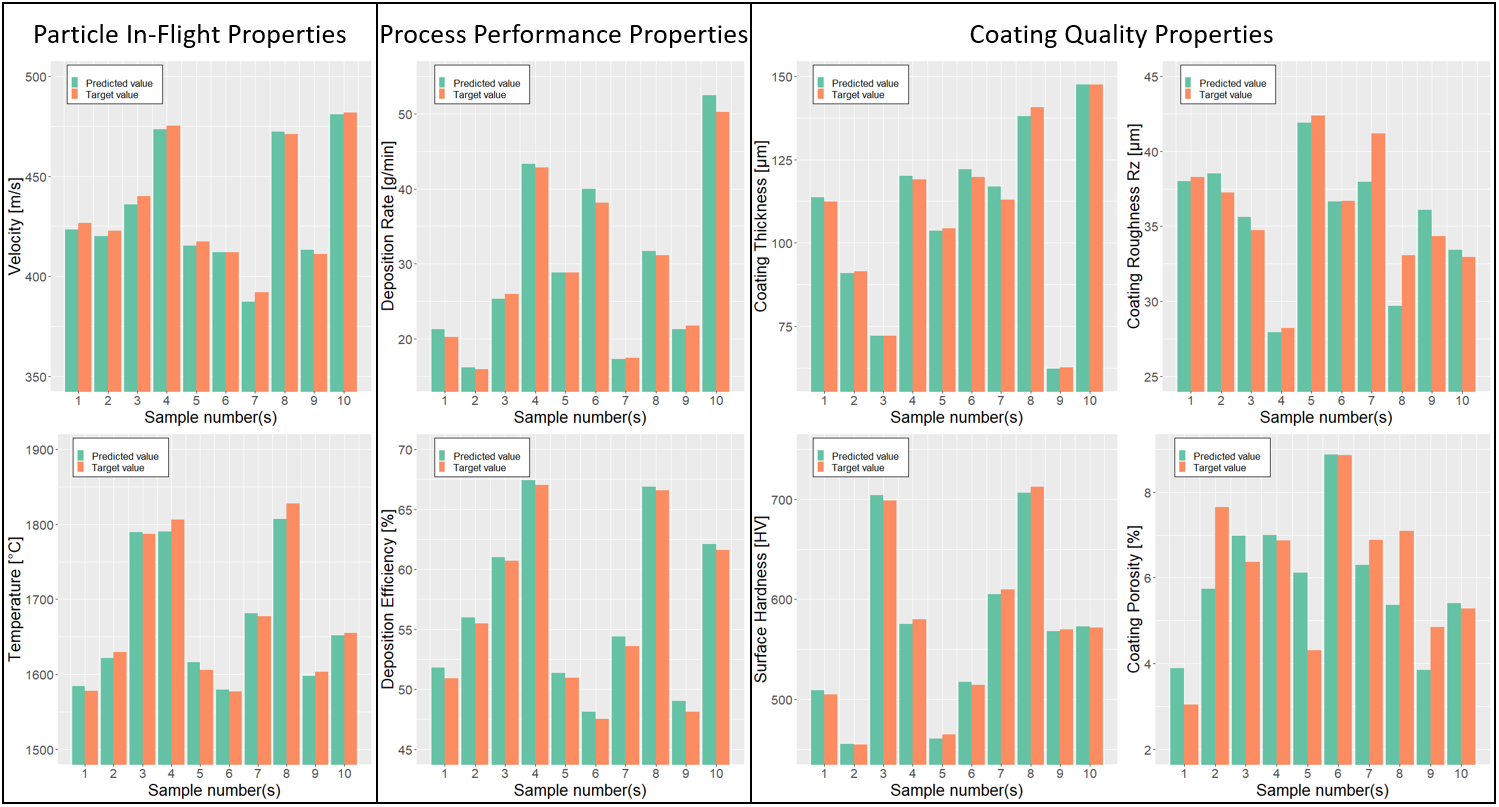}}
\caption{Prediction performance comparison on test data. The bar charts depict the predicted (green) and measured (orange) values of TCCSM properties across ten test samples.}
\label{fig:TEST_RESULTS}
\end{figure*}

Following the analysis conducted on particle in-flight velocity, the other targets falling under the PIP, PPP, and CQP classifications are treated in a similar manner. The hyperparameters $(p, C)$ for each model are carefully selected by applying the described n-fold cross-validation. This procedure ensures maximum performance across the entire spectrum of quality parameters. After training and validation, the models were tested using the test data provided, which consists of another ten labeled samples. Importantly, this test set was used solely for evaluating the model's generalization performance on previously unseen data and was not involved in the fitting or tuning process. Figure \ref{fig:TEST_RESULTS} summarizes the performance of the models on this test data for the prediction of TCCSM relevant properties. The figure depicts predicted values (in green) alongside measured values (in orange) across eight sub-graphs displayed as bar charts. The x-axis represents sample numbers (1-10), while the y-axis, truncated to improve presentation clarity, denotes the corresponding quality properties categorized according to Table \ref{tab:properties}. The trained SEMKL models show a high suitability for prediction, with minimal errors in the prediction of PIP and PPP. 

Furthermore, the quality predictor exhibits strong performance in predicting coating thickness and surface hardness. However, slight discrepancies between predicted and actual values are evident for coating roughness and coating porosity, particularly for samples 7, 8, and 2, 5, 8, 9, respectively. However, the discrepancies observed in coating porosity measurements are likely influenced by the volatile nature of the measurement technique using Image Analysis (IA), as discussed in \cite{ref26}.

The accuracy of the developed SEMKL models was further assessed by calculating both RMSD and $\ell_\epsilon$ error for the test data, as shown in Table \ref{tab:comparisionLM}. While it is important to note that the linear model is trained using squared loss, whereas SEMKL is optimized with $\ell_\epsilon$ loss, the SEMKL model significantly outperformed the classical linear model, included as a baseline for comparison, across all eight HVOF properties. The optimized SEMKL models, utilizing hyperparameters $p$ and $C$ tuned for minimal error, demonstrated consistent predictive accuracy. Notably, the RMSD values for the SEMKL model are closely aligned across both the test set and the training/validation stages, indicating proper generalization despite the relatively small test set of ten samples. As mentioned earlier, RMSD provides a more precise basis for comparison, as evident in deposition efficiency where both models perform well; however, SEMKL predictions align more closely with target values, highlighting its superior fit. Despite minor discrepancies in coating roughness and porosity for certain samples, the overall performance remains satisfactory, confirming the suitability of SEMKL for predictive modelling in thermal spray processes. Addressing these nuanced variations could further improve the predictive accuracy of the proposed framework and may warrant attention in future studies.

\begin{table}[ht]
\centering 
\footnotesize
\begin{tabular}{l l l r r r r}  
\toprule\toprule
\multicolumn{1}{c}{} & \multicolumn{4}{c}{SEMKL} & \multicolumn{2}{c}{Linear Model} \\
\cmidrule(r){2-5} \cmidrule(r){6-7}
 Properties & $p$ & $C$ & RMSD & $\ell_\epsilon$ error & RMSD & $\ell_\epsilon$ error \\
\midrule
Particle in-flight velocity & $2^{10}$ & $10^4$ & 7.028 & 60.848 & 19.871 & 186.523  \\
Particle in-flight temperature  & $2^{10}$ & $10^3$ & 18.073 & 151.524 & 33.072 & 319.275 \\
Deposition rate & $2^{9}$ & $10^2$ & 1.104 & 5.641 & 2.729 & 20.199 \\
Deposition efficiency & $2^{10}$ & $10^2$ & 0.014 & 0 & 0.037 & 0 \\
Coating thickness & $2^{9}$ & $10^1$ & 3.212 & 24.187 & 11.037 & 101.033 \\
Coating roughness & $2^{10}$ & $10^0$ & 2.234 & 18.105 & 6.210 & 58.881 \\
Coating hardness & $2^{10}$ & $10^0$ & 8.954 & 72.031 & 15.720 & 147.497 \\
Coating porosity & $2^{9}$ & $10^2$ & 1.242 & 6.947 & 3.226 & 26.024 \\
\bottomrule\bottomrule
\end{tabular}
    \caption{Comparison of RMSD and \(\ell_\epsilon\) error values for various coating and process properties in the HVOF process, modeled using SEMKL and a classical linear model. The SEMKL models are reported with the hyperparameters $p$ and $C$, optimized during model selection to minimize prediction error.}
    \label{tab:comparisionLM}
\end{table}

Additionally, the practical model implementation of the developed SEMKL models and the complementary cooperation with the data aggregator allows for proactive notification to the spray booth operators in real-time, when deviations from pre-defined quality standards are expected. Based on this notification, operators can promptly adjust the process to ensure consistent adherence to quality standards. 

To further demonstrate the practical integration of the developed framework and its real-time capabilities, the following illustrations provide a comprehensive view of the overall process pipeline. These representations highlight both the physical operation of the coating system and the interaction between the data-driven predictive models and the on-site operators. By capturing these critical aspects, the illustrations emphasize how real-time analytics and predictions are employed to maintain process consistency and adherence to predefined quality parameters in industrial-scale thermal spraying applications.

\begin{figure}[!ht]
\centerline{\includegraphics[width=\textwidth, height = 8cm]{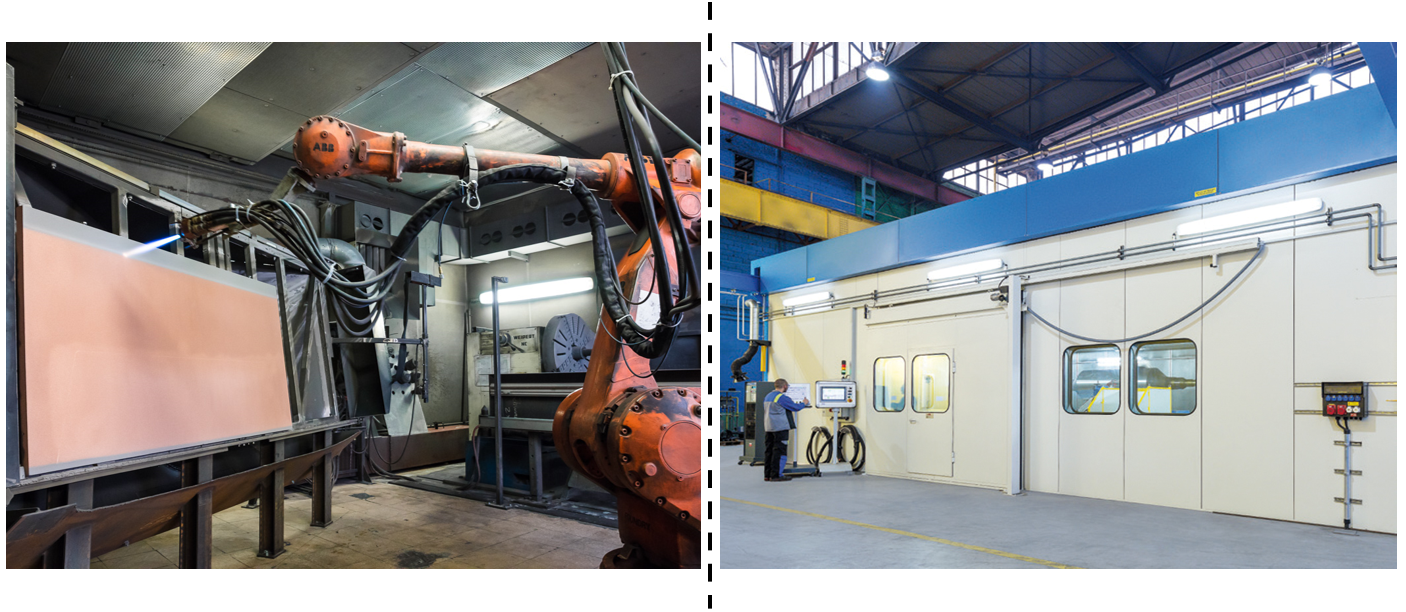}}
\caption{Visual representation of the thermal spray process for TCCSM coating. The left panel shows the application of a thermal spray coating on a large-scale plate by a robotic system within the spray booth at voestalpine Stahl GmbH \cite{TSM2}. The right panel depicts operator interaction with the system controller next to the spray booth, where the validated framework is operational.}
\label{fig:booth}
\end{figure}

Figure~\ref{fig:booth} provides a visual representation of the TCCSM coating process setup. On the left, the thermal spraying of a large-scale plate, used in the steel manufacturing process, is shown. The plate acts as an example of a TCCSM and is the base material for the thermal coating, which is applied to improve wear resistance under high thermal and mechanical stress. On the right, the thermal spray booth is depicted, with an operator interacting with the system controller. This illustrates the operator's role in monitoring and adjusting the process parameters based on system feedback. The booth is designed to manage the high-velocity combustion process, ensuring that critical operational conditions are maintained throughout the coating application.

\begin{figure}[!ht]
\centerline{\includegraphics[width=\textwidth, height = 9cm]{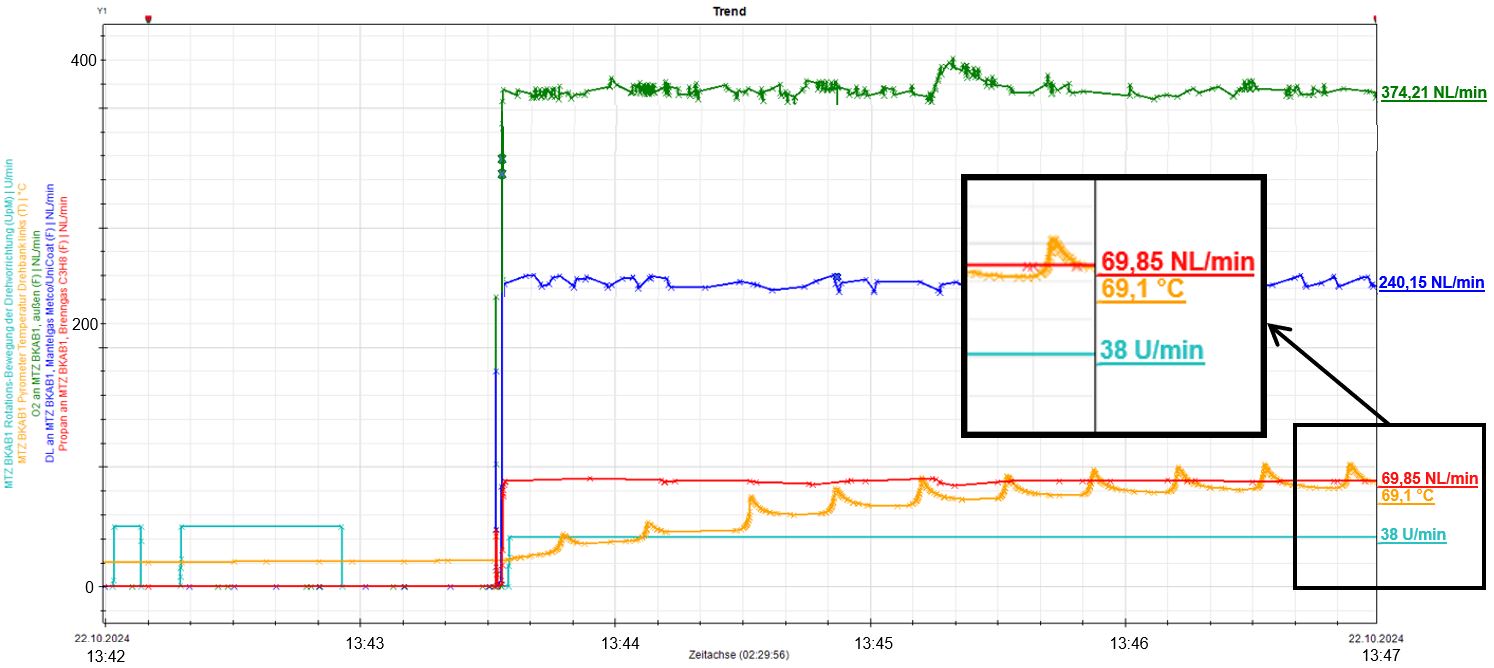}}
\caption{Interface of the web application within the validated framework for real-time monitoring of the thermal spray process, displaying live data on critical process variables, including propane, compressed air, and oxygen flow rates, pyrometer temperature readings, and lathe rotational speed, with an average display latency of 4.92~ms.}
\label{fig:MEdisLive}
\end{figure}

Following this overview of the coating process setup and operator interactions, Figure~\ref{fig:MEdisLive} further illustrates the real-time data processing capabilities of the system, highlighting a selection of the critical media involved in the coating operation and their monitoring through the data aggregator’s web application. This figure displays the interface of the web application, providing live data on propane and compressed air flow rates, oxygen flow, and the temperature measured by the pyrometer on the surface of the coated TCCSM. Additionally, it depicts the current rotational speed of the turning lathe, with an average time delay of 4.92 ms between event detection and the display of process variables, as discussed in Section \ref{sect:agg}. For enhanced visualization, the live values of three key components are presented in an enlarged window, facilitating immediate assessment and adjustment by the operators.

\begin{figure}[!ht]
\centerline{\includegraphics[width=\textwidth]{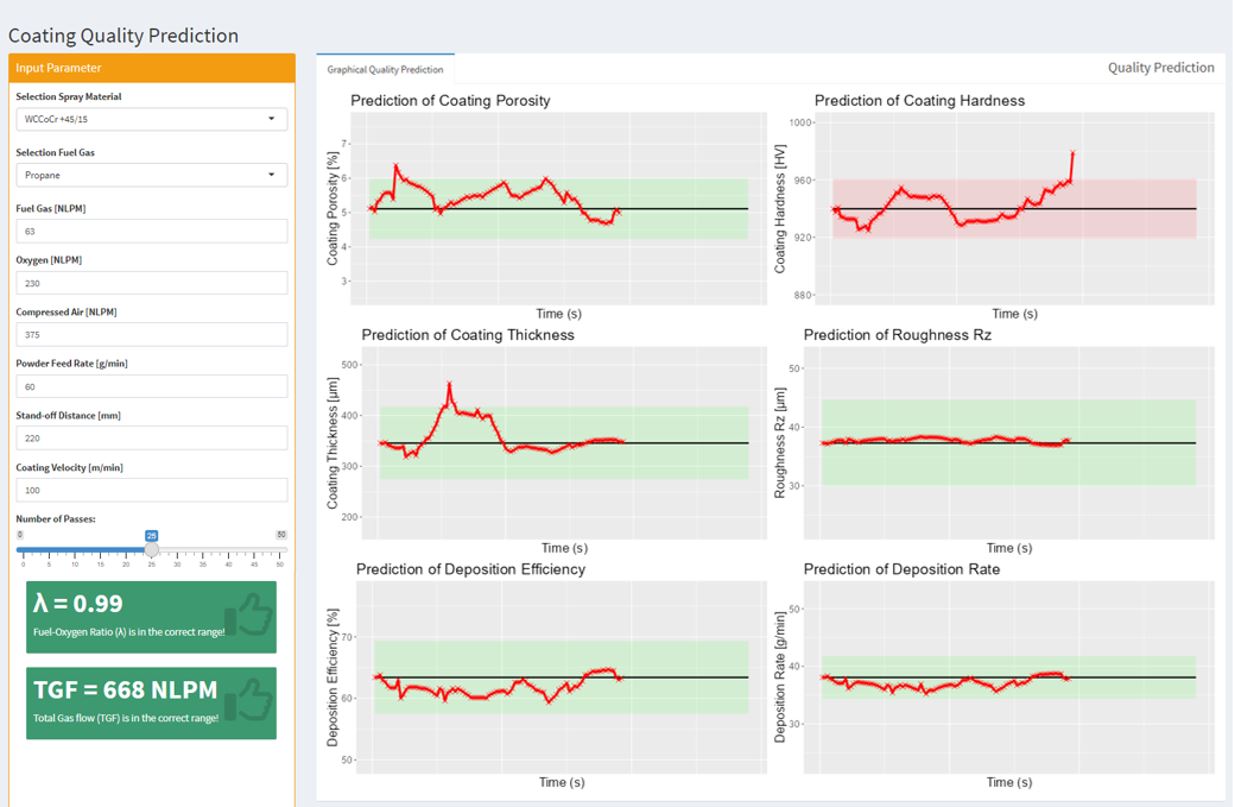}}
\caption{Real-time prediction of six coating characteristics--porosity, hardness, thickness, roughness, deposition efficiency, and deposition rate--using the quality predictor, displayed alongside target input parameters and live calculations.}
\label{fig:hvofapp}
\end{figure}

Building on the data processing capabilities of the data aggregator in Figure~\ref{fig:MEdisLive}, Figure~\ref{fig:hvofapp} demonstrates the integration of the quality predictor for coating quality assessment. On the left side of the interface, an extract from the target input parameters is displayed, while the lower section presents live-calculated key process metrics, such as the fuel-oxygen ratio. The central panel features real-time predictions from the quality predictor for six selected coating properties: coating porosity, coating hardness, coating thickness, coating roughness, deposition efficiency, and deposition rate. These line charts continuously update as the coating process progresses, graphically indicating whether the predicted values remain within predefined quality limits, depending on the TCCSM intended application.

As shown in Figure~\ref{fig:hvofapp}, five of the six properties currently remain within acceptable boundaries, while coating hardness is trending outside the set limits. Although the system does not offer automated corrective actions, operators can use the real-time monitoring features of the web application (Figure~\ref{fig:MEdisLive}) to investigate the associated media inputs and adjust the process parameters accordingly. Ongoing collaborative efforts with voestalpine Stahl GmbH aim to refine the system to optimize input parameters to consistently achieve multiple targeted coating characteristics. Additionally, this collaboration includes the development of automated corrective mechanisms that not only detect deviations but also recommend parameter adjustments, such as modifying fuel flow rates or adjusting spray distance, to maintain coating quality within specified limits.

Furthermore, the latency of the system--defined as the maximum time between receiving a trigger event, performing the prediction analysis via the quality predictor, and displaying the results--has been measured at between 6,150 ms and 6,830 ms. This latency range confirms that the predictive model operates efficiently within a real-time framework, providing operators with sufficient time to respond to deviations and take corrective action to maintain the desired coating quality standards.

While the real-time framework demonstrates the effective application of SEMKL for coating quality prediction, it is important to consider the challenges associated with scaling this approach in industrial environments. One potential limitation is the computational complexity associated with training and deploying SEMKL models on vast amounts of real-time data generated in industrial settings. This complexity may hinder scalability, as the computational resources required for processing and analyzing data from numerous sensors across multiple production lines can be substantial. Additionally, SEMKL's adaptability may be affected by the need for continuous model updates and retraining to accommodate evolving process dynamics and changing operating conditions in large-scale industrial environments.

\section{Conclusion}

The standardized coating process for TCCSM represents a considerable risk for the quality of innovative steel products. This article proposes a novel solution involving real-time data analytics and predictive quality management to upgrade established processes in TCCSM maintenance. By combining continuous process monitoring and a data-driven methodology, two principal components, i.e., the data aggregator and the quality predictor, were designed. Toward the goal of predictive quality management, the quality predictor, powered by SEMKL has shown its prominent ability to predict TCCSM relevant quality properties based on pre-processed real-time data. Meanwhile, the data aggregator, equipped with sensors and intelligent data handling, provides operators with real-time insights into the coating process and promptly notifies them of any deviations from expected quality standards, facilitating continuous process monitoring.

The solution presented in this article offers a new direction for practitioners in the field of maintenance utilizing thermal spray technology. In addressing the complexities of real-time data in manufacturing settings, a combination of advanced pre-processing techniques was implemented to optimize the quality of data inputs for the SEMKL model. Specifically, techniques such as outlier detection and removal, temporal alignment, and noise reduction were found to be particularly effective in enhancing data quality. Moreover, leveraging the domain knowledge of thermal spray technicians allowed for a focus on technical features relevant to the coating process, thereby further boosting the prediction accuracy of the applied SEMKL model.

The application of kernel methods such as SEMKL is proven to be powerful in realizing predictive quality management. Further research activities are encouraged to jointly improve the advanced maintenance strategy of the TCCSM. In addition to real-time prediction and notification capabilities, the integration of the proposed framework with the data aggregator provides enhanced data-driven decision-making processes. This synergy allows stakeholders to gain valuable insights into process dynamics, optimize resource allocation, and proactively address potential quality deviations, thus fostering continuous improvement and operational excellence in TCCSM maintenance processes. 

Although our study has not explicitly examined the scalability of our SEMKL-based predictive system across different production scales and operational conditions within the steel industry, its potential applicability to various production scales is noteworthy. Exploring its adaptability to different operational contexts could provide valuable insights into its versatility and effectiveness in predictive maintenance practices in steel manufacturing.

\section*{Declarations}

\subsection*{Acknowledgements}
    The authors gratefully acknowledge voestalpine Stahl GmbH for their support through the research center, provision of materials, and financial contribution to this investigation. 

\subsection*{Funding}
    This research was funded in part by the Austrian Science Fund (FWF) SFB 10.55776/F68 ``Tomography Across the Scales'', project F6805-N36 (Tomography in Astronomy). For open access purposes, the author has applied a CC BY public copyright license to any author-accepted manuscript version arising from this submission.

\subsection*{Availability of data and materials}
    The datasets generated and/or analyzed during the case study are not publicly available due to company confidentiality but are available from the corresponding author on reasonable request. The data used in this manuscript are proprietary and subject to confidentiality agreements. However, interested parties may request access to the data by contacting the corresponding author. Requests will be considered on a case-by-case basis, subject to company compliance and confidentiality agreements.

\subsection*{Competing interests}
     The authors declare that they have no competing interests.

\subsection*{Authors' contributions}
    WR conceived and designed the study, collected the data, conducted the analysis, estimation, and modelling, and drafted the manuscript. The case study was carried out by WR and CH. CH also contributed to data interpretation and offered technical explanation and revisions. SH and RR supported the study with theoretical insights and provided critical feedback. All authors reviewed and approved the final manuscript.    
\clearpage

\printglossary[type=\acronymtype]
\clearpage
\bibliographystyle{plain}
{\footnotesize
\bibliography{mybib}
}

\end{document}